\documentclass[]{article}
\usepackage{xcolor}
\usepackage{standalone}
\usepackage{tikz}
\usepackage{tikz}
\usepackage[margin=1in]{geometry}
\usepackage{mathpazo}

\usetikzlibrary{arrows.meta,positioning,fit,calc}

\usepackage[colorinlistoftodos]{todonotes}

\begin{document}

\title{Towards Efficient and Reliable AI Through Neuromorphic Principles}
\author{
    Bipin Rajendran\thanks{Corresponding author: bipin.rajendran@kcl.ac.uk},
    Osvaldo Simeone,
    Bashir M. Al-Hashimi \\
    \\
    Centre for Intelligent Information Processing Systems, \\
    Department of Engineering, King’s College London, \\
    WC2R 2LS, United Kingdom
}

\date{} 

\newcommand{\subject}[1]{}
\newcommand{\keywords}[1]{}

\subject{Artificial intelligence, machine learning, neuromorphic computing, hardware lottery, uncertainty quantification}

\keywords{efficient AI, reliable AI, neuromorphic computing, spiking neural networks, continual learning, in-memory computing}







\maketitle

\begin{abstract}
Artificial intelligence (AI) research today is largely driven by ever-larger neural network models trained on graphics processing units (GPUs). This paradigm has yielded remarkable progress, but it also risks {entrenching a hardware lottery} in which algorithmic choices succeed primarily because they align with current hardware, rather than because they are inherently superior. In particular, the dominance of Transformer architectures running on GPU clusters has led to an arms race of scaling up models, resulting in exorbitant computational costs and energy usage. At the same time, today’s AI models often remain unreliable in the sense that they cannot properly quantify uncertainty in their decisions -- for example, large language models tend to hallucinate incorrect outputs with high confidence.

This article argues that achieving more {efficient and reliable} AI will require embracing a set of principles that are well-aligned with the goals of  {neuromorphic engineering}, which are in turn  inspired by how the brain processes information. Specifically, we outline six key neuromorphic principles, spanning algorithms, architectures, and hardware, that can inform the design of future AI systems: (\emph{i}) the use of {stateful, recurrent models}; (\emph{ii}) extreme {dynamic sparsity}, possibly down to spike-based processing; (\emph{iii}) {backpropagation-free on-device learning} and fine-tuning; (\emph{iv}) {probabilistic decision-making}; (\emph{v}) {in-memory computing}; and (\emph{vi}) hardware-software co-design via {stochastic computing}. We discuss each of these principles in turn, surveying relevant prior work and pointing to directions for research. 
\end{abstract}
\section{Introduction}
The current dominant paradigm in AI centers on training very large neural networks through highly parallel computing on GPUs or TPUs. This approach has yielded impressive capabilities, from foundation models that can be adapted to myriad tasks \cite{bommasani2021opportunities} to generative systems for text and multi-modal data. However, the paradigm is beginning to show signs of strain. The computational resources required to push state-of-the-art performance are growing exponentially, as a single training run of a top-tier language model can consume millions of GPU-hours of compute and megawatt-hours of energy \cite{touvron2023llama}. Such heavy demands raise sustainability concerns and create an uneven playing field where only a few industry actors can afford to participate \cite{nyt_ai_quartz_2025}. 

At the same time, current design methodologies leave important gaps in terms of reliability, despite the use of Internet-scale datasets. Models often do not know when they don’t know: they can be confidently wrong, failing to provide calibrated uncertainty estimates \cite{guo2017calibration,zecchin2023robust}. This has practical consequences: for instance, a conversational agent may output a plausible-sounding but false statement  with unwarranted certainty, or an autonomous system may not recognize novel conditions under which its predictions become unreliable.

Aggravating both the efficiency and reliability challenges is a lack of diversity in the algorithms and hardware being explored. Due to 
the ``hardware lottery,'' research ideas that align well with existing hardware  and software ecosystems have thrived, whereas alternative approaches are often unexplored and held to an excessively high standard \cite{Hooker2020}. The Transformer architecture is a case in point -- it rose to prominence in part because its pattern of dense matrix multiplications maps efficiently onto GPUs, enabling rapid scaling. As specialized hardware and infrastructure continue to be built around Transformers, it becomes increasingly difficult for fundamentally different models or learning approaches to gain traction. 

This article starts from the observation that  insights from neuroscience and neuromorphic engineering  have started to be integrated into some AI solutions and implementations, pointing to a new possible path for AI.  The human brain achieves remarkable intelligence with a power budget of about 20~W, suggesting there are organizational principles very different from those of power-hungry digital processors. Neuromorphic engineering builds on results from neuroscience, which tells us that brain computation is event-driven, sparse, and inherently uncertain \cite{barrett2020seven}. Specifically, biological neurons communicate via discrete spikes, operate via recurrent, stateful, mechanisms, and continuously adapt through local synaptic plasticity, all the while seamlessly integrating memory and processing in the same computational substrates. 

As mentioned, recent work  has begun to embrace these insights. For example, the DeepSeek language model uses a Mixture-of-Experts (MoE) design to activate only a small fraction of its 671B parameters for any given query, dramatically reducing computation and cost \cite{deepseek2024report}. Recent state-of-the-art models such as FP8-trained Transformers and INT8-inference LLMs operate with activations as few as {8 bits} \cite{micikevicius2022fp8,dettmers2022llmint8,xiao2022smoothquant}, and research demonstrates reliable {4-bit} activation quantization  \cite{esser2019lsq}, with specialized binary/ternary networks pushing activations to {1–2 bits} \cite{rastegari2016xnor,zhou2016dorefa,wei2023bitnet}. Furthermore, breakthroughs in implementation and model design have demonstrated recurrent and state-space architectures with performance matching -- and in some settings surpassing -- Transformers, while enhancing efficiency  \cite{gu2023mamba,sun2023retnet,peng2023rwkv,gu2022efficient}.
 
  More broadly, in this article, we identify six neuromorphic principles that can guide this shift:  
\begin{itemize}
\item \textbf{Stateful, recurrent models:} Incorporating temporal state and feedback loops allows networks to retain context and process sequential information efficiently, reducing the need for the all-to-all or autoregressive attention mechanisms of Transformers, whose complexity scales quadratically with the context size. This mirrors the brain’s continuous, stateful, integration of information over time (see Fig. \ref{fig:transformer-vs-recurrent}).  

\item \textbf{Dynamic sparsity via  discrete activations:} Event-driven and sparse activation schemes minimize computation and energy use by engaging only the necessary submodels,  neurons, or synapses at each step. Spiking networks and conditional computation mechanisms such as MoE  embody this principle, aligning computation cost with informational relevance (see Fig. \ref{fig:chartsparse}).  

\item \textbf{Backpropagation-free on-device learning.} Local learning rules with limited feedback do not require backpropagation and can vastly reduce memory requirements for training or fine-tuning, enabling continual on-device adaptation (see Fig. \ref{Fig:mezobo}). 

\item \textbf{Probabilistic decision-making.} Representing and sampling uncertainty allows AI systems to provide calibrated confidence estimates and robust decisions under ambiguity. Techniques such as Bayesian inference and best-of-$N$ sampling align with how biological systems reason under uncertainty (see Fig. \ref{fig:3}).  

\item \textbf{In-memory computing.} Performing computation directly where data are stored eliminates costly data movement and supports massively parallel operations. Analog   resistive memory arrays exemplify this principle, offering orders-of-magnitude gains in energy efficiency for neural workloads (see Fig. \ref{Fig:arch}).  

\item \textbf{Stochastic computing and hardware–software co-design.} Embracing device-level noise as a computational resource enables efficient sampling, optimization, and uncertainty estimation. This principle highlights the synergy between physical randomness and probabilistic algorithms, leading to co-designed systems that are both efficient and inherently reliable (see Fig. \ref{Fig:accece}).  
\end{itemize}

In the rest of this review-style article, we discuss each of these principles in turn, surveying relevant prior work and providing an outlook on the road ahead for neuromorphic-inspired AI.

\section{Stateful and Recurrent Neural Processing}

Artificial neural networks were originally inspired by biological neurons \cite{mcculloch1943logical, hebb1949organization}, yet most mainstream models today are fundamentally \emph{stateless} mappings from inputs to outputs. In a standard deep feed-forward network, each neuron’s activation is a memoryless function of the current input. Transformers, which have largely supplanted recurrent architectures in sequence modeling, substitute this intrinsic memory with self-attention mechanisms and positional encodings that allow for context aggregation over sequences \cite{Vaswani17}. While highly parallelizable, these models are inherently static at inference time and require computation of attention weights over the entire past sequence at each step during inference (see left panel of Fig. \ref{fig:transformer-vs-recurrent}).

\begin{figure*}[t]
\centering
\resizebox{\linewidth}{!}{%
 \begin{tikzpicture}[x=1cm,y=1cm,>=Latex,font=\footnotesize]
\tikzset{
  layer/.style={draw,rounded corners=3pt,minimum width=5.8cm,minimum height=0.9cm,fill=black!05,align=center},
  token/.style={draw,rounded corners=2pt,minimum width=0.9cm,minimum height=0.55cm,fill=white},
  state/.style={draw,circle,minimum size=7mm,fill=black!05},
  arrow/.style={-Latex,thick,shorten >=0.5pt,shorten <=0.5pt},
}

\node[font=\bfseries] at (0,4.8) {Transformer (Stateless)};

\node[token] (x1) at (-2.8,3.6) {$x_1$};
\node[token] (x2) at (-1.4,3.6) {$x_2$};
\node[token] (x3) at ( 0.0,3.6) {$x_3$};
\node[token] (x4) at ( 1.4,3.6) {$x_4$};
\node at ( 2.1,3.6) {$\cdots$};
\node[token] (x5) at ( 2.8,3.6) {$x_T$};

\node[layer] (attn1) at (0,2.2) {Self-Attention + MLP};
\node[layer] (attn2) at (0,0.8) {Self-Attention + MLP};
\node[layer] (attn3) at (0,-0.6) {Self-Attention + MLP};

\foreach \t in {x1,x2,x3,x4,x5}
  \draw[arrow] (\t.south) -- ($(\t.south |- attn1.north)$);

\foreach \t in {x1,x2,x3,x4,x5} {
  \draw[arrow] ($(\t.south |- attn1.south)$) -- ($(\t.south |- attn2.north)$);
  \draw[arrow] ($(\t.south |- attn2.south)$) -- ($(\t.south |- attn3.north)$);
}

\node[token] (y1) at (-2.8,-2.0) {$y_1$};
\node[token] (y2) at (-1.4,-2.0) {$y_2$};
\node[token] (y3) at ( 0.0,-2.0) {$y_3$};
\node[token] (y4) at ( 1.4,-2.0) {$y_4$};
\node at ( 2.1,-2.0) {$\cdots$};
\node[token] (y5) at ( 2.8,-2.0) {$y_T$};

\foreach \t in {y1,y2,y3,y4,y5}
  \draw[arrow] ($(\t.north |- attn3.south)$) -- (\t.north);

\begin{scope}[xshift=10.5cm]
\node[font=\bfseries] at (0,4.8) {Recurrent (Stateful)};

\node[token] (xr1) at (-2.8,3.6) {$x_1$};
\node[token] (xr2) at (-1.4,3.6) {$x_2$};
\node[token] (xr3) at ( 0.0,3.6) {$x_3$};
\node[token] (xr4) at ( 1.4,3.6) {$x_4$};
\node at ( 2.1,3.6) {$\cdots$};
\node[token] (xr5) at ( 2.8,3.6) {$x_T$};

\node[state] (h1) at (-2.8,2.2) {$h_1$};
\node[state] (h2) at (-1.4,2.2) {$h_2$};
\node[state] (h3) at ( 0.0,2.2) {$h_3$};
\node[state] (h4) at ( 1.4,2.2) {$h_4$};
\node at ( 2.1,2.2) {$\cdots$};
\node[state] (h5) at ( 2.8,2.2) {$h_T$};

\foreach \i/\n in {1/xr1,2/xr2,3/xr3,4/xr4,5/xr5}
  \draw[arrow] (\n.south) -- (h\i.north);

\draw[arrow] (h1.east) -- (h2.west);
\draw[arrow] (h2.east) -- (h3.west);
\draw[arrow] (h3.east) -- (h4.west);
\draw[arrow] (h4.east) -- (h5.west);

\node[token] (yr1) at (-2.8,0.7) {$y_1$};
\node[token] (yr2) at (-1.4,0.7) {$y_2$};
\node[token] (yr3) at ( 0.0,0.7) {$y_3$};
\node[token] (yr4) at ( 1.4,0.7) {$y_4$};
\node at ( 2.1,0.7) {$\cdots$};
\node[token] (yr5) at ( 2.8,0.7) {$y_T$};
\foreach \i in {1,2,3,4,5} \draw[arrow] (h\i.south) -- (yr\i.north);
\end{scope}

\node[font=\small] at (5.25,4.8) {};

\end{tikzpicture}

}
\caption{Architectural comparison between transformer and recurrent sequence processing paradigms. (Left) Transformer (stateless): All input tokens are processed simultaneously in parallel through a stack of identical layers, each consisting of self-attention mechanisms followed by feedforward networks (MLPs).  The computation is stateless, allowing for highly parallelizable training and inference. (Right) Recurrent models (stateful): Input tokens are processed sequentially, with each token combined with the previous hidden state to produce the current hidden state through a recurrent transition function. Information propagates through the sequence via explicit state-to-state connections (horizontal arrows), creating a temporal dependency chain. This stateful processing enables constant memory overhead regardless of sequence length, but introduces sequential bottlenecks that limit parallelization.}
\label{fig:transformer-vs-recurrent}
\end{figure*}

By contrast, biological neurons are \emph{stateful}: their membrane potentials integrate inputs over time, and their spiking activity depends on this evolving internal state \cite{gerstner2002spiking}. This endows even a single neuron with short-term memory and nonlinear temporal dynamics. \emph{Recurrent Neural Networks} (RNNs) were an early attempt to incorporate such statefulness into artificial systems \cite{elman1990finding}. In RNNs, the hidden state evolves as a function of both the current input and the previous state, thereby introducing a form of discrete-time recurrence (see right panel of Fig.  \ref{fig:transformer-vs-recurrent}). Variants such as Long Short-Term Memory (LSTM) networks \cite{hochreiter1997long} and Gated Recurrent Units (GRUs) \cite{cho2014learning} demonstrated the practical utility of this approach in sequential domains including speech recognition \cite{graves2013speech} and language modeling \cite{mikolov2010recurrent}.

However, training recurrent networks through backpropagation through time (BPTT) proved challenging due to issues of vanishing and exploding gradients \cite{bengio1994learning}, and their inherently sequential computation prevented efficient parallelization on GPUs. These limitations contributed to the widespread adoption of Transformer models, which replace explicit recurrence with multi-head self-attention \cite{Vaswani17}. Yet, this paradigm shift came with a cost: Transformers effectively ``unroll'' recurrence into depth, replicating the same processing across many layers. This approach achieves global context modeling but at the expense of quadratic computational complexity in the length of the context \cite{tay2020efficient}.

Recent research has revisited the idea of stateful sequence processing through alternative formulations that preserve parallelism while reintroducing recurrence. One major development is the class of \emph{state-space models} (SSMs), which learn latent dynamical systems with input-dependent gating \cite{gu2022efficient, gupta2022diagonal}. SSMs maintain a compact internal state that evolves linearly in time, modulated by learned gates, yielding efficient long-range dependency modeling. Importantly, their recurrence can be expressed in a convolutional form that allows for fully parallel training on GPUs, overcoming the classical bottlenecks of RNNs. Variants such as S4 \cite{gu2022efficient} and Mamba \cite{gu2023mamba} exemplify this paradigm, offering the computational efficiency of convolutional models while preserving the inductive bias of temporal state evolution.

\emph{xLSTM}  is a modern recurrent architecture that generalizes the classical LSTM by introducing multiplicative gating mechanisms inspired by SSMs. It is reported to achieve performance competitive with transformers on long-context benchmarks \cite{voelker2024xlstm}, but training cannot be parallelized \cite{danieli2025}. Parallel training can be restored via methods based on the recurrent solution of non-linear equations via linearization steps \cite{danieli2025}.

Closely related are models based on \emph{linear attention}, which obviate the quadratic complexity of self-attention via a form of recurrence based on a matrix state evaluated over aggregated key-value states \cite{katharopoulos2020Transformers, choromanski2021rethinking}. Another promising direction involves hybrid architectures such as \emph{Retentive Networks} (RetNets) \cite{sun2023retnet} and \emph{RWKV models} \cite{peng2023rwkv}. Both approaches operates on the same associative memory principle of Transformers \cite{behrouz2025s}, thus operating on the basis of  key, queries, and values, but they  effectively linearize the temporal dependency structure of Transformers,  capturing long-range context with subquadratic complexity. Similar mappings between Transformer's keys, queries, and values and SSM operations have been also put forth \cite{bick2024transformers}.

RNNs, linear attention, RetNets, and RWKV all iteratively update a hidden state, represented as vectors or matrices in a given space. A similar principle is applied by \emph{looped Transformer models}, which apply the same multi-head attention model a number of times
 to transform a given hidden set of tokens \cite{dehghani2018universal}. The number of applications of the looped layer can vary depending on the input, and each separate recursive step may be viewed as a "thought" in a reasoning sequence \cite{zhu2025scaling}.

Neuromorphic computing platforms natively support stateful computation. Chips such as Intel’s Loihi \cite{davies2018loihi} implement spiking neurons as dynamical systems that continuously evolve in time, making them natural substrates for recurrent models. Hybrid approaches that integrate spiking dynamics with continuous-time SSMs have been recently proposed for energy-efficient event-driven learning \cite{yin2023ssmspiking}.

In summary, incorporating stateful recurrent mechanisms into deep architectures enables native and efficient processing of temporal data streams. Rather than unrolling computations across many feed-forward layers, these models maintain persistent internal states that evolve over time. When coupled with neuromorphic or asynchronous hardware, such architectures offer the potential for always-on, low-power processing of sensory and sequential data with extended memory horizons.

\section{Dynamic Sparsity and Event-Driven Spiking}

The principle of \emph{dynamic sparsity} posits that an intelligent system should recruit only the minimal necessary set of resources -- neurons, synapses, or computations -- at any given time as a function of the current stimuli. Biological brains exemplify this: at any instant, only a small fraction of neurons are active \cite{barrett2020seven,buzsaki2019brain}. Sparse activation enhances metabolic efficiency and increases the robustness and capacity of neural representations \cite{olshausen2004sparse,barlow1961possible,attwell2001energy}. Translating this concept to artificial systems, dynamic sparsity implies that, instead of activating every neuron in a layer for each input, as in dense feedforward or Transformer networks, the model selectively activates a subset of neurons or weights conditioned on the input \cite{evci2020rigging}.

\begin{figure*}[t]
\centering
\includegraphics[width=\linewidth]{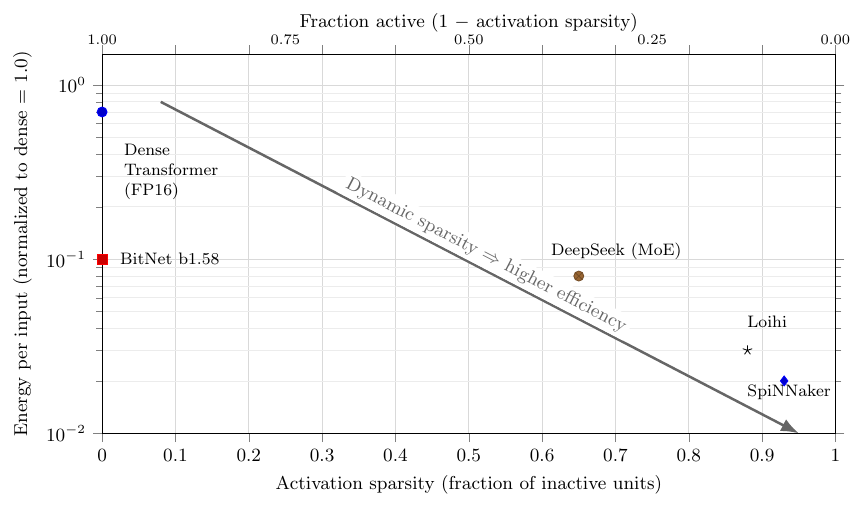}
\caption{Energy efficiency versus dynamic sparsity for various neural network architectures. Dyanamic sparsity refers to the fraction of computational units (neurons or network components) that produce zero or negligible output for a given input, effectively remaining inactive during inference. The y-axis shows energy consumption per input normalized to a dense FP16 transformer baseline. The figure illustrates distinct approaches to achieve computational efficiency: BitNet b1.58 achieves energy reduction through extreme quantization, while not leveraging dynamic sparsity; DeepSeek (MoE) employs mixture-of-experts routing that activates a fraction of the model per input; Neuromorphic hardware platforms, such as Loihi and SpiNNaker, leverage event-driven spiking neural networks, attaining sparsity at the level of individual neurons. The diagonal arrow indicates the general trend: architectures that exploit dynamic, input-dependent sparsity tend to achieve substantially higher energy efficiency, approaching the extreme efficiency of biological neural systems.}
\label{fig:chartsparse}
\end{figure*}

\emph{Spiking neural networks} (SNNs) -- the canonical model in neuromorphic computation -- embody dynamic sparsity at the level of neural activation \cite{maass1997networks,roy2019towards}. In SNNs, neuron outputs are discrete \emph{spike events} rather than continuous-valued activations. In a well-designed system, spikes occur  infrequently, and if a neuron does not spike, its synapses remain idle, consuming virtually no dynamic power \cite{davies2018loihi,furber2014spinnaker}. 

Consequently, in neuromorphic chips, energy use scales roughly with the number of spikes rather than the total number of synapses or neurons \cite{davies2018loihi,roy2019towards}. In particular, when a spike is transmitted, the postsynaptic neuron need only add the corresponding synaptic weight to its internal state, thus avoiding costly floating-point multiplications. In variants where spikes carry integer-valued payloads, multiplications reduce to low-cost bit shifts and additions \cite{wu2025neuromorphic}. Recent large-scale AI models have begun to integrate similar ideas. For instance, {BitNet b1.58} \cite{wang2024bitnet} operates with ternary $(-1, 0, +1)$ weights and 8-bit integer activations.

Neuromorphic hardware is inherently well suited to exploit sparse and event-driven computation. Conventional accelerators such as GPUs and TPUs, struggle with fine-grained irregular sparsity, since they must mask or skip blocks of operations -- wasting bandwidth and control cycles \cite{han2016eie}. In contrast, neuromorphic processors perform computation only when triggered by spike events  \cite{davies2018loihi}. 

A complementary approach to dynamic sparsity is \emph{conditional computation} or \emph{gating}, in which only a subset of parameters or subnetworks is used for a given input. MoE models exemplify this paradigm \cite{shazeer2017outrageously,fedus2022switch}. A gating network selects one or a few experts to process each token, skipping the rest. DeepSeek’s architecture \cite{deepseek2024report} combines MoE with multi-head latent attention, achieving up to a 20$\times$ reduction in active operations relative to dense models of comparable quality.

In summary, dynamic sparsity and event-driven processing align computation with actual demand, minimizing the waste inherent in always-on dense architectures. These principles apply across scales -- from individual neurons in SNNs to modular experts in MoE systems -- and can be further amplified by hardware that supports fine-grained power gating.   Using the energy per input -- per token in the case of an LLM -- of a full-precision dense Transformer as reference, Fig. \ref{fig:chartsparse} shows quantized models such as BitNet b1.58 requiring a fraction, say 0.1–0.5$\times$, of the energy due the use of fewer bits per multiply-accumulate operation, while not increasing sparsity.  MoE models may activate a small fraction of MLP experts, increasing sparsity and decreasing energy consumption. Neuromorphic chips, such as Loihi or SpiNNaker  typically reports orders-of-magnitudes  better energy efficiency via sparsity.


\section{Backpropagation-Free On-Device  Learning}

\begin{figure}[!ht]
\centering
\includegraphics[width=1\textwidth]{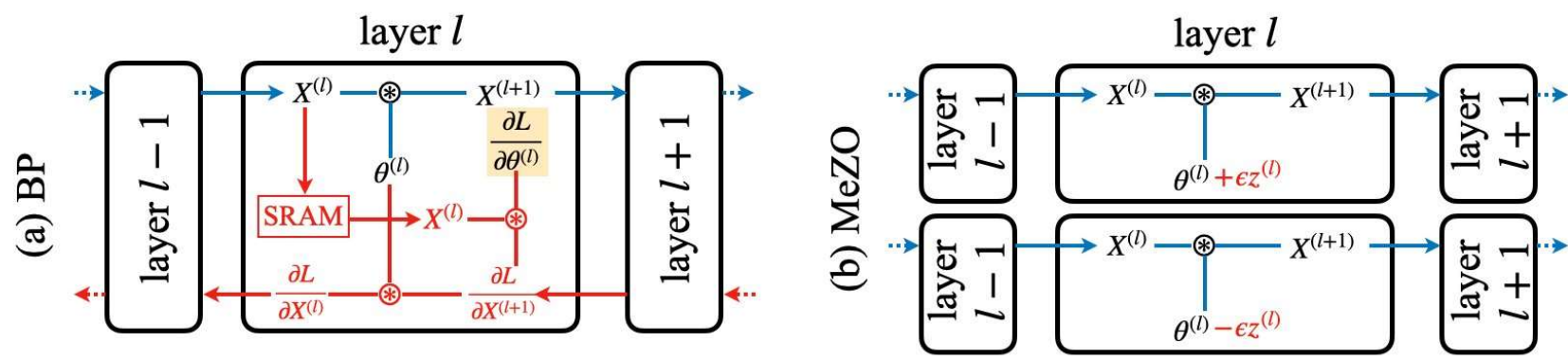}
\caption{While backprop (BP) requires the storage of all the activations produced in the forward pass, imposing a hard constraint on the model sizes that can be stored within an on-device memory, memory-efficient ZO (MeZO) optimization only requires forward passes.}\label{Fig:mezobo}
\end{figure}

Training deep networks with backpropagation, or backprop, imposes severe memory demands, primarily due to the need to store intermediate activations during the forward pass so that gradients can later be computed in reverse (see Fig. \ref{Fig:mezobo}(a)). For a Transformer model with 
$L$ layers, hidden dimension 
$D$, batch size $B$, and context length $N$, 
 the total activation storage scales with the product  
$BLND$ \cite{korthikanti2023activation, malladi2023mezo}. In practice, this means that the memory required for fine-tuning often exceeds that of inference. Even techniques such as activation checkpointing only partially mitigate this problem, reducing but not eliminating the need to retain a subset of activations in memory. Consequently, backprop-based fine-tuning may be impractical on memory-constrained or on-device hardware, such as neuromorphic processors or edge accelerators, unless the context is sufficiently small (see \cite{song2025memory}).

Neuromorphic and edge-AI systems highlight this tension between learning capability and memory efficiency. On such devices, both model weights and activations must fit entirely within on-chip memory, excluding the possibility of offloading to external DRAM. This constraint severely limits the deployable model size under backpropagation. For instance, fine-tuning a Transformer model may require up to 12 times the memory required for inference, and thus   even an A100 GPU with 80GB memory can accommodate only about a 2.7 billion parameter model during backpropagation-based fine-tuning with Adam \cite{malladi2023mezo}. In such settings, the fundamental bottleneck is not computation, but the storage of intermediate states.

A growing body of work has explored backpropagation-free learning schemes that remove this dependency on stored activations. Among these, \emph{zeroth-order (ZO) optimization} offers a principled and hardware-friendly alternative. Instead of computing gradients via explicit backpropagation, ZO methods estimate them through function evaluations,  typically by perturbing the model parameters in random directions and observing the corresponding change in loss \cite{spall1998spsa, duchi2015zo} (see Fig. \ref{Fig:mezobo}(b)).

Modern implementations, such as Memory-Efficient Zeroth-Order (MeZO) optimization, require only forward passes, making their memory footprint essentially equivalent to that of inference \cite{malladi2023mezo}. This enables models up to three times larger to fit on the same hardware budget, with potential gains exceeding 100$\times$ for long-context models \cite{malladi2023mezo}. While ZO methods traditionally suffered from slow convergence, these recent advances leverage the structure of pre-trained models to constrain the effective optimization space, allowing competitive fine-tuning performance on large models \cite{liu2024sparsemezo, yang2024adazeta}.

Because ZO training eliminates backward passes and stored activations, it aligns naturally with the \emph{local learning rules} that are  natively implemented on neuromorphic architectures \cite{davies2018loihi,jang19:spm}. Each forward evaluation can be mapped to local circuit dynamics, and parameter perturbations can be realized as lightweight analog modulations rather than large-scale digital gradient computations. 

More broadly, neuromorphic principles suggest  learning mechanisms that are local, incremental, and continual. A hallmark of such mechanisms is the absence of a separate backward pass with precise error signals propagated through the exact transposed network. The resulting local rules  use only information available at a synapse, namely  the activities of the pre- and post-synaptic neurons, and possibly a diffuse modulatory signal indicating reward or novelty.  The modulatory reward signal can take the form of \emph{direct feedback alignment}, whereby the backward weights are not tied to the forward weights, but are random projections  from the output layer to hidden layers \cite{nokland2016direct}. Conceptually, MeZO-like methods are also well aligned with the goals of continual learning due to their effective optimization of a smoothed-out learning objective \cite{yu2025memory}.

Geoffrey Hinton’s \emph{Forward-Forward algorithm} offers an alternative completely backprop-free approach. This approach trains networks by presenting positive (real data) and negative (fake or corrupted data) examples and adjusting weights to make the network’s layer activations ``good'' (high) for positives and ``bad'' (low) for negatives, using only forward passes \cite{Hinton2022forward}.

\begin{figure}[!t]
\centering
\includegraphics[width=0.8\textwidth]{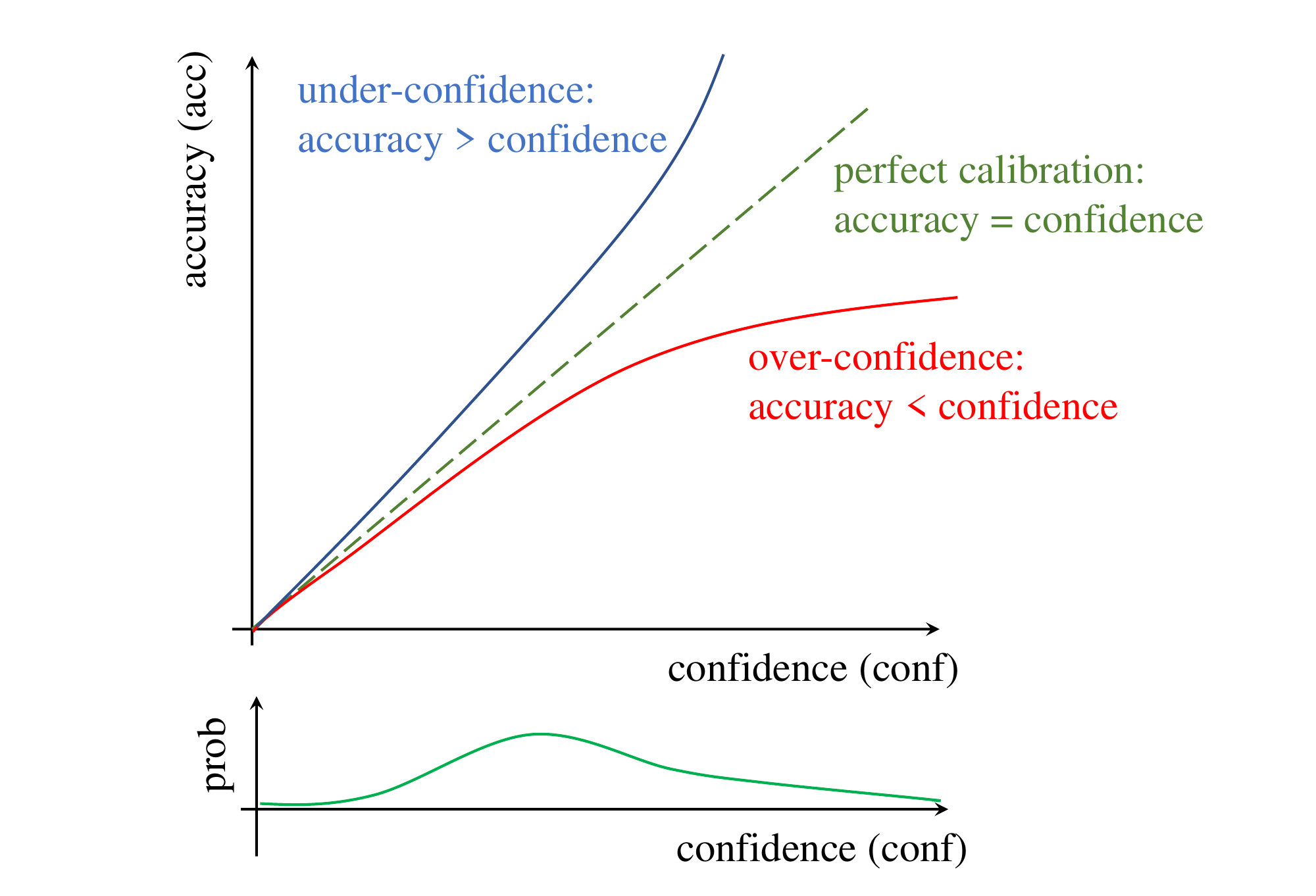}
\caption{Calibration of an AI model is typically measured via a \emph{reliability diagram}. As shown in the top panel, a reliability diagram plots the true accuracy (measured on a test set) versus the confidence level that the model assigns to its decisions. The true accuracy is estimated by evaluating the average accuracy of all decisions made with a given confidence level. It is also useful to plot a histogram of the confidence levels produced by the model (bottom panel), which allows one to visualize the distribution of confidence and identify biases (e.g., tendency to rarely predict low-confidence outputs).}\label{fig:3}
\end{figure}

\section{Probabilistic Decision-Making and Uncertainty Quantification}

For AI systems to be trustworthy in real-world settings, they must be able to assess and communicate their own uncertainty. A reliable AI should not only provide predictions or actions, but also an indication of confidence or a distribution over possible outcomes. In traditional deep learning, a model often produces a ``hard'' output, such as a class label, or a softmax probability distribution that is treated as a measure of confidence. However, these probabilities are frequently uncalibrated, meaning that the model’s 90\% confidence might correspond to only a 70\% empirical success rate in practice, a common symptom of overconfidence \cite{guo2017calibration} (see Fig.~\ref{fig:3}). 

Conventional methods to quantify uncertainty are based on different forms of \emph{ensembling}. Ensembling may take place at several levels -- model, input (prompt), or output -- and the general principle is to estimate uncertainty from the degree of disagreement between predictions obtained from different ensemble members.

At the \emph{model level}, multiple models (or multiple variations of the same model) are trained independently, and their predictions are aggregated. If all models produce similar outputs, confidence is high; if they diverge, this signals uncertainty. Deep ensembles, which average the predictions of independently trained neural networks, have been shown to significantly improve both calibration and predictive accuracy \cite{lakshminarayanan2017simple}. 

A principled probabilistic framework that underlies model ensembles is provided by \emph{Bayesian approaches} to learning \cite{simeone2022book}.  In a Bayesian framework, the model parameters are treated as random variables, and the model outputs are averaged over the posterior distribution of these parameters. Exact Bayesian inference is generally intractable for modern neural networks, leading to approximate methods such as Monte Carlo (MC) sampling and Variational Inference (VI). MC methods, such as MC dropout, approximate uncertainty by sampling from stochastic networks at test time \cite{gal2016dropout}, while VI replaces the true posterior with a tractable approximate distribution that is optimized during training \cite{blundell2015weight}. 

Beyond model-level variation, ensembling can also take place at the level of the \emph{input}. In large language models, \emph{prompt ensembling} generates diverse outputs by varying the prompt or query formulation while keeping the model fixed. This technique leverages the stochasticity of autoregressive generation: different prompt phrasings can yield slightly different reasoning trajectories or responses. Aggregating these outputs -- through majority voting or confidence-weighted combination -- provides an estimate of epistemic uncertainty and improves calibration \cite{jiang2023calibrating}. Such methods are particularly appealing because they require no retraining and are readily applicable to large pretrained models.

At the \emph{output level}, ensembling often takes the form of test-time sampling strategies such as \emph{Best-of-$N$} decoding. Instead of accepting a single forward pass, one draws multiple outputs -- by sampling random seeds, adding small perturbations to inputs or weights, or varying temperature parameters -- and then selects the best candidate according to an external metric, such as a learned reward model or a heuristic for coherence or consistency. In generative models, this strategy yields higher-quality results and provides an implicit estimate of confidence based on the diversity of candidate outputs \cite{nan2025can}.

Although ensemble-based methods are empirically effective, they generally lack theoretical guarantees on calibration or coverage. In contrast, \emph{conformal prediction} offers a distribution-free framework that provides formal uncertainty bounds without requiring strong assumptions on the underlying model or data distribution \cite{vovk2005algorithmic,angelopoulos2023conformal}. The core idea is to assign a nonconformity score to each prediction and construct a prediction region that is guaranteed, under mild conditions, to contain the true outcome with a specified probability. Conformal prediction thus converts uncalibrated model outputs into calibrated prediction sets or intervals, offering rigorous finite-sample guarantees. Recent work  has extended conformal prediction to modern learning settings, including neural networks, federated systems, and sequential decision-making (see, e.g., \cite{zecchin2024forking,zhu2023federated,simeone2025conformal}).

\section{In-Memory Computing for AI}

The separation of memory and processing in conventional computing architectures -- commonly referred to as the \emph{von Neumann bottleneck} -- has become a critical limitation for AI workloads. As neural networks grow in size and depth, the energy and latency associated with moving weights and activations between memory and the processor can dominate the total computation cost \cite{horowitz2014energy}. 
For example, a multiply-and-accumulate (MAC) operation, which is the fundamental computation involved in neural networks, consumes less than 10 femtojoules (fJ) when implemented in advanced CMOS technology, whereas the energy cost of accessing data -- even from on-chip SRAM -- can be 100 to 1000 times higher \cite{Murmann2021}. This imbalance makes the traditional approach of repeatedly fetching weights from off-chip DRAM fundamentally inefficient for large-scale AI inference and training.

\emph{In-memory computing} (IMC) aims to mitigate this problem by performing computation directly where the data reside -- typically inside or adjacent to the memory array, minimizing data movement \cite{ielmini2018memory, sebastian2020memory}. The key idea is to exploit the physical properties of memory devices to implement the MAC operation within the memory substrate itself.  In such crossbars, the internal atomic configuration of the memory cell is modulated by applying electrical pulses such that its effective conductance can be programmed in an anlog manner to encode a synaptic weight. Voltage signals encoding the elements of the input vector are applied along the rows of the crossbar, and  the resulting currents, proportional to the product of input voltage and cell conductance, flow along the columns and are summed by Kirchhoff’s current law. The total current in each column thus represents the analog weighted sum of inputs -- a physical implementation of a MAC operation in a single step. This architectural concept has been experimentally validated on a variety of emerging memory technologies such as phase change memory (PCM), resistive random access memory (ReRAM) and spin-transfer torque random access memory (STTRAM) devices \cite{ Khaddam2022, Jung2022, Wen2024}; see \cite{Aguirre2024} for a review. 

Alongside analog IMC, there is active research in fully digital \emph{processing-in-memory} (PIM) architectures, which integrate logic operations directly into SRAM or DRAM arrays \cite{JiHoon2021,SukhanLee2021,Donghyuk2022} or modifying the SRAM cell architecture as well as  the peripheral circuits used for reading to implement MAC operation directly within the array \cite{Mingu2014,ChuanJia2021}. These digital approaches retain full numerical precision and are more compatible with existing CMOS design flows.  While digital PIM/IMC offers less energy gain than analog IMC, it provides higher reliability and programmability, making it a promising near-term path for commercial deployment.


The benefits of analog IMC are twofold: massive inherent parallelism and a drastic reduction in data movement. All weights remain stationary in the crossbar,  greatly reducing data transfer overhead. Furthermore, the small footprint of analog memory devices allows very high compute density, and the absence of digital multipliers leads to substantial energy savings. Experimental results indicate that analog or CMOS-analog hybrid IMC architectures can achieve energy efficiencies 10–100$\times$ greater than equivalent digital implementations \cite{Aguirre2024}.

\begin{figure}[!ht]
\centering
\includegraphics[width=0.85\textwidth]{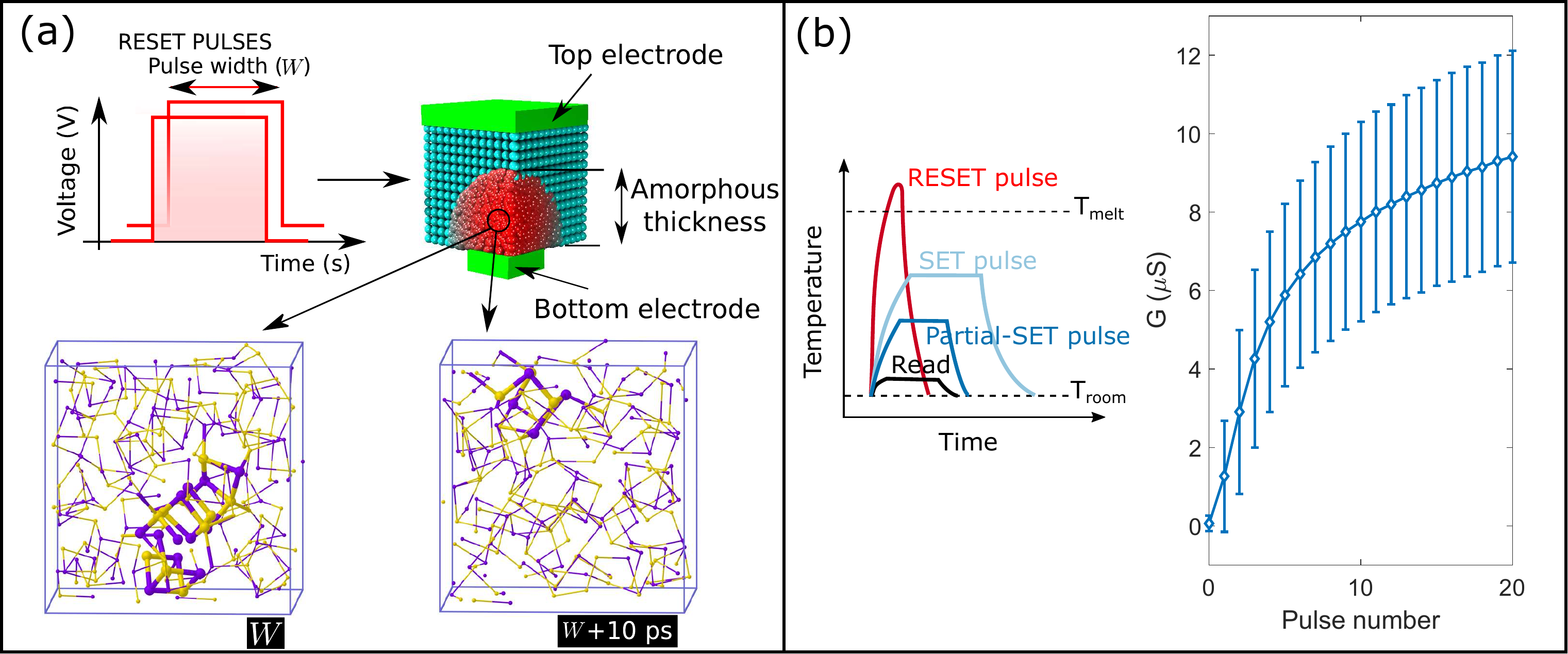}
\caption{Programming noise observed in nanoscale phase change memory devices. (a)~Molecular dynamics simulations show that even slight variations in the programming pulse conditions result in vastly different atomic configurations in chalcogenide materials; figure adapted from \cite{Gallo2016}. (b)~Sequential application of partial-SET programming pulses (left) results in stochastic conductance distributions for a PCM device (right); figure adapted from \cite{Rajendran48}.}\label{pcmnoise}
\end{figure}

However, these gains come with the challenge of \emph{imprecision and noise}. Resistive devices suffer from device-to-device variability, stochastic switching behavior, and conductance drift over time \cite{Rajendran64} (see Fig.~\ref{pcmnoise}). Peripheral circuits add further sources of error, such as thermal noise, parasitic capacitances, and limited dynamic range \cite{LeGallo2023}. As a result, a straightforward mapping of a deep neural network to an analog crossbar typically incurs severe accuracy degradation compared to a digital baseline \cite{Burr14}.

To address these issues, researchers have proposed several complementary strategies. One is \emph{hardware-aware training}, in which the network is trained or fine-tuned while injecting noise or quantization errors that mimic hardware nonidealities, thereby improving robustness to device variability \cite{Joshi2020}. Another is hardware-level redundancy and statistical averaging, such as the “committee machine” architecture proposed in \cite{joksas2020committee}, which uses multiple replicated networks whose outputs are averaged to reduce the impact of random device errors. Hybrid \emph{analog–digital} architectures have also been proposed, especially for implementing on-chip learning, where matrix multiplications are performed in analog using memristive cross-bar arrays, while gradients are accumulated in high precision using dedicated digital circuits, with the memristive weights updated periodically based on the accumulated gradients \cite{Rajendran76,Wen2024}.


While in-memory computing seeks to mitigate the data movement bottleneck by modifying the memory architecture to execute MAC operations in place, neuromorphic computing takes this integration further by mimicking the parallel and adaptive nature of biological neural systems. Noting the similarity between the characteristics of conduction through biological ion channels and sub-threshold transport in MOS transistors, Carver Mead pioneered analog electronic circuits that mimic the dynamics of neurons and synapses in the mid-1980s, laying the foundations for the field of \emph{neuromorphic engineering} \cite{meadbook}. This approach was prevalent for over two decades, during which hardware prototypes with increasing complexity and functionality were demonstrated, using spike-based realizations of computation, memory, learning, and communication. However, these implementations were limited in scale (in terms of network size), since they were designed using transistors from older technology nodes and were constrained by challenges associated with controlling, debugging, and automating complex designs based on analog electronics. 

Later, purely digital CMOS-based neuromorphic hardware prototypes began to be developed, with prominent examples including SpiNNaker (University of Manchester) \cite{SpiNNaker}, TrueNorth (IBM) \cite{truenorth}, and Loihi (Intel) \cite{davies2018loihi}. Leveraging the advances of Moore’s law scaling, these chips and systems have achieved hardware neural networks with over 1~million neurons, albeit at significant energy cost compared to equivalent biological systems;  see \cite{Rajendran84} for a review.


Over the last decade, significant research efforts have also been directed at developing custom nanoscale electronic devices that mimic key computational features of neurons and synapses using memristive materials. Such non-CMOS-based realizations are attractive for their potential power- and area-efficiency, and recent works have demonstrated proof-of-concept integration of such nanoscale devices with mainstream CMOS technology \cite{Greatorex2025}.




\section{Stochastic Computing with Physical Noise}
An alternative paradigm for AI hardware design that is being pursued today moves away from the traditional bifurcated focus on (i) training algorithm development and (ii) post-training mitigation of nanoscale device noise for hardware implementation. Instead, it advocates a co-optimisation approach that harnesses nanoscale device noise and stochasticity \emph{as a computational resource} to realize uncertainty-aware on-hardware inference and learning algorithms. Some prominent examples include leveraging programming and read noise of PCM devices (Fig.~\ref{pcmnoise}) to implement Bayesian binary spiking neural networks \cite{Rajendran87} and exploiting intrinsic memristor variability for in situ learning through Markov Chain Monte Carlo sampling to achieve probabilistic inference directly within hardware \cite{dalgaty2021situ}. Similar implementations have been demonstrated using other memristive devices, including RRAM  \cite{Gao2023FaultTolerant, Bonnet2023}, {Ferroelectric FET} \cite{Dutta2022},  {Ferroelectric NAND}\cite{Song2025},  {Fe-diode} \cite{Huang2025} and spintronic devices \cite{Tuhin2023}, enabling efficient realisations of Bayesian ML architectures that provide uncertainty quantification based on intrinsic hardware stochasticity.

\begin{figure}[!ht]
\centering
\includegraphics[width=0.55\textwidth]{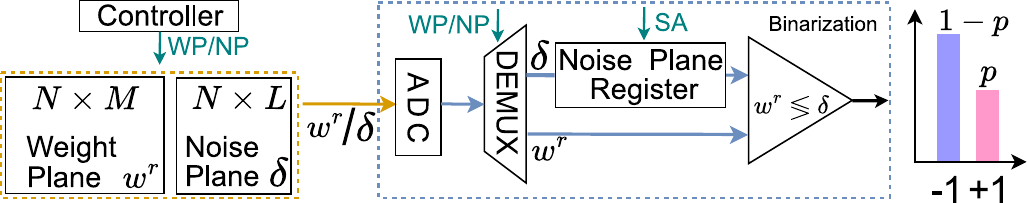}\hfill
\includegraphics[width=0.45\textwidth]{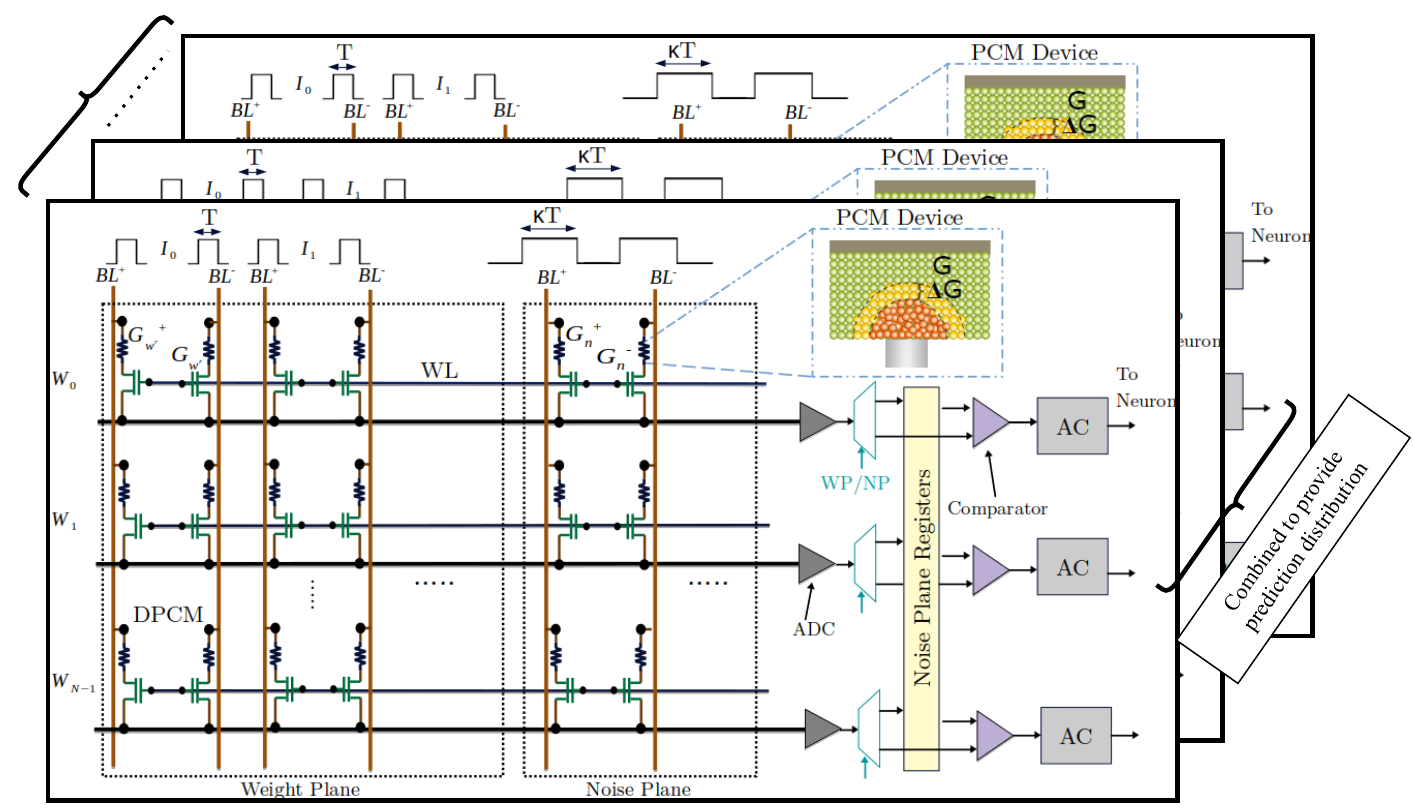}
\caption{The PCM-based crossbar architecture introduced in \cite{Rajendran87}, which leverages the inherent stochasticity of nanoscale analog hardware to represent uncertainty. Figure adapted from \cite{Rajendran87}.}\label{Fig:arch}
\end{figure}

A key feature in these implementations is the exploitation of intrinsic device noise and stochasticity as a resource for Bayesian sampling, obviating the need for expensive pseudo-random number generator circuits implemented using CMOS linear feedback shift registers (LFSR). For example, reference \cite{Rajendran87} demonstrated a Bayesian neuromorphic system trained using the approach described in \cite{jang2021bisnn}. The random weights in this system -- representing the learner’s uncertainty due to limited data -- are stored in hardware as the conductances of nanoscale PCM devices organized in crossbar arrays. Accordingly, the required randomness is generated across an ensemble of PCM differential cells using the devices’ inherent stochasticity (as illustrated in Fig.~\ref{Fig:arch}). 

\begin{figure}[!ht]
\centering
\includegraphics[width=0.65\textwidth]{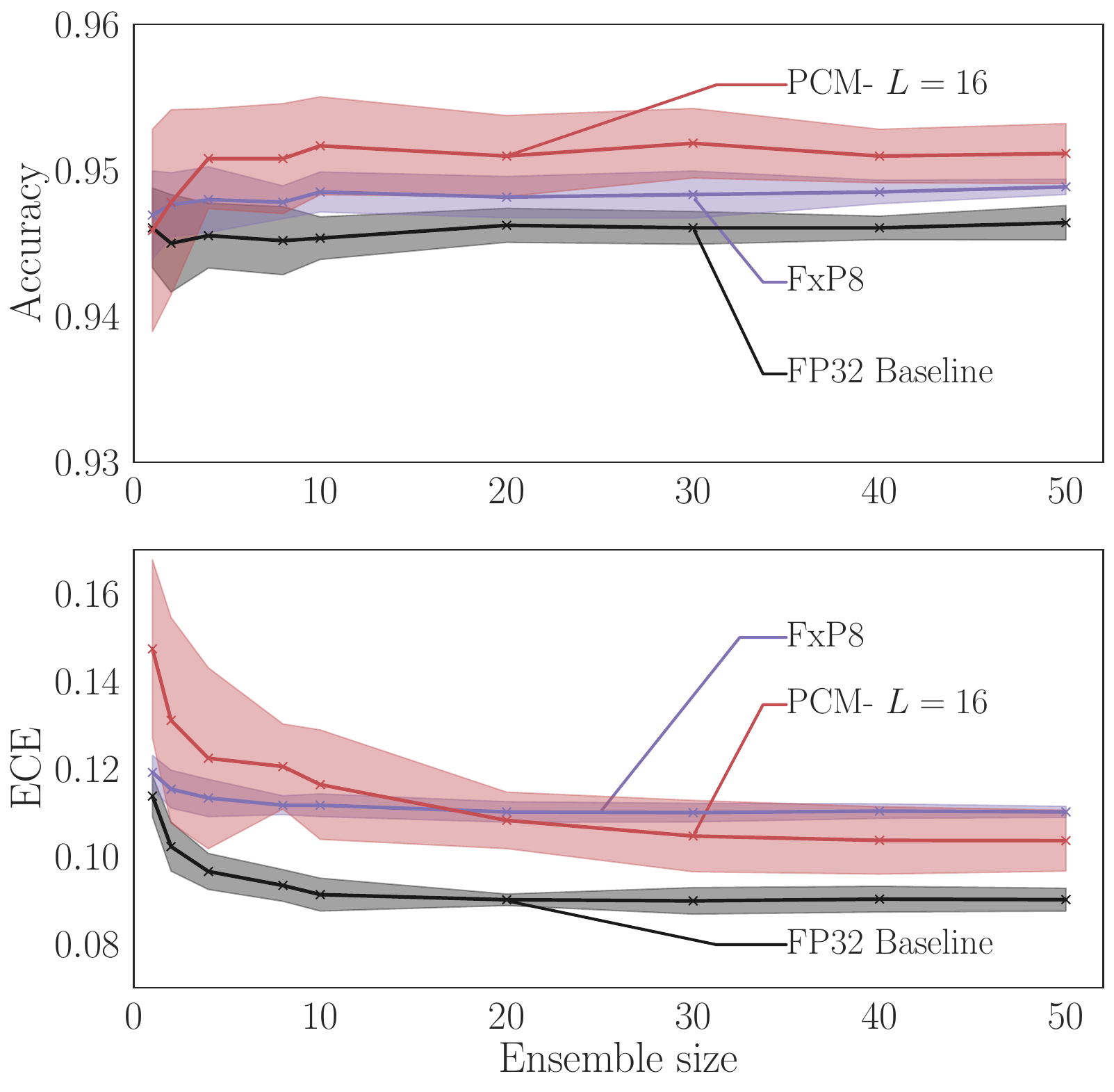}
\caption{Accuracy and Expected Calibration Error (ECE) for the Wisconsin Breast Cancer dataset, based on simulations of the architecture proposed in \cite{Rajendran87} using nanoscale PCM device variability. Results are benchmarked against ideal software simulation (32-bit floating point, FP32) and a reduced-precision fixed-point hardware implementation (8-bit, FxP8). Figure adapted from \cite{Rajendran87}.}\label{Fig:accece}
\end{figure}

Based on transistor-level estimates, reference \cite{Rajendran87} reports that the PCM-based core design is over $9\times$ more area-efficient than an equivalent SRAM crossbar implementation with 8-bit fixed-point weights -- chosen to ensure a fair CMOS benchmark while maintaining comparable accuracy and precision. Overall, the architecture in \cite{Rajendran87} demonstrates how nanoscale device variability can be exploited within a Bayesian framework to enable uncertainty-aware, trustworthy decision-making, while consuming significantly fewer hardware resources.  (see Fig.~\ref{Fig:accece}).
 A related promising direction that is being explored is thermodynamic computing, where the inherent thermal noise of physical systems has been leveraged to implement computational tasks such as Gaussian sampling and matrix inversion \cite{melanson2025thermodynamic}.

\section{Conclusions}
AI systems have become extraordinarily large, expensive, and power-hungry, yet they remain brittle in the face of uncertainty. In this article, we have advocated for an alternative path forward --  one that draws inspiration from neuromorphic principles to design AI systems that are both efficient and reliable. We identified six key principles, namely stateful recurrent processing, dynamic sparsity, backpropagation-free learning, probabilistic decision-making, in-memory computing, and stochastic physical computation, and we discussed how each of these can address the limitations of the current paradigm.

A unifying theme across these principles is the idea of \emph{hardware-algorithm co-design}: by rethinking algorithms with an eye toward how they will be implemented on physical devices, and vice versa, developing devices tailored to the needs of advanced algorithms, we can unlock capabilities that neither conventional software nor hardware alone could achieve. For instance, treating noise as a feature allows hardware to perform random sampling intrinsically, enabling efficient uncertainty quantification. 

We are already seeing promising results from efforts aligning AI research with these six principles. Research groups and companies are exploring alternative computing paradigms such as event-driven vision sensors paired with spiking processors for ultra-low-power perception, on-device learning for personalization, and analog accelerators for running large language model inference at a fraction of the energy cost of digital accelerators. These proof-of-concept systems demonstrate significant gains in either efficiency or reliability, or both. 

Looking ahead, achieving the full potential of efficient and reliable AI through neuromorphic principles will require interdisciplinary collaboration. Advances in materials science and device engineering are needed to create better memory devices, sensors, and stochastic primitives for hardware. At the same time, breakthroughs in algorithms and circuits will be crucial to effectively harness the intrinsic behavioral characteristics of these devices. The intersection of information theory, neuroscience, and machine learning is envisaged here to provide theoretical foundations for understanding how to optimize these new systems. Examples of specific research topics include hardware implementations of zero-th order fine-tuning on device for models with dynamic sparsity, the fusion of local learning rules with zero-th order optimization, and the development of effective and efficient Bayesian Transformers and SSMs. Future research may also investigate possible synergies with probabilistic processing enabled by quantum computers  (see, e.g., \cite{chen2025stochastic}).

The coming years are likely to witness rapid advancements in this space, and we hope this article serves as a roadmap and inspiration for researchers and engineers to jointly explore the rich opportunities at the intersection of neuroscience, hardware, and AI.

\bibliographystyle{IEEEtran}
\bibliography{refs.bib}
\end{document}